\documentclass{llncs}
\usepackage[utf8]{inputenc}
\usepackage{amsmath}
\usepackage{amssymb}
\usepackage{mathtools}
\usepackage[table,dvipsnames]{xcolor}
\usepackage{siunitx}
\usepackage[ruled,vlined,linesnumbered,noend]{algorithm2e}

\usepackage[margin=0.3cm]{caption}
\usepackage{subcaption}
\usepackage{booktabs}
\usepackage{wrapfig}
\usepackage{multirow}

\usepackage{pifont}
\newcommand{\cmark}{\ding{51}}%
\newcommand{\xmark}{\ding{55}}%

\usepackage{scalerel}

\newcommand{\sectref}[1]{Section~\ref{#1}}

\newcommand{\figref}[1]{Figure~\ref{#1}}
\newcommand{\tabref}[1]{Table~\ref{#1}}

\newcommand{\defref}[1]{Definition~\ref{#1}}

{\bfseries}{\itshape}
\newcounter{exampcount}
\newcommand{\startpara}[1]{{%
\vskip6pt\noindent
{\bf #1.}}}

\def\matr#1{{\mathbf{#1}}}

\def\Nset{\mathbb{N}}
\def\Zset{\mathbb{Z}}
\def\Iset{\mathbb{I}}

\def\Rset{\mathbb{R}}

\renewcommand{\leq}{\leqslant}
\renewcommand{\geq}{\geqslant}

\def\squareforqed{\hbox{\rlap{$\sqcap$}$\sqcup$}}
\def\qed{\ifmmode\squareforqed\else{\unskip\nobreak\hfil
\penalty50\hskip1em\null\nobreak\hfil\squareforqed
\parfillskip=0pt\finalhyphendemerits=0\endgraf}\fi}

\newcommand\dist{{\mathit{Dist}}}
\newcommand\Dist{{\dist}}
\newcommand\Path{{\mathit{Path}}}

\renewcommand{\Pr}[2]{{\mathit{Pr}_{#1}^{#2}}}

\def\AP{{\mathit{AP}}}

\def\Act{{A}}
\def\act{a}

\def\Real{\mathbb{R}}
\newcommand\inner[2]{\langle #1,#2 \rangle}
\DeclareMathOperator{\supp}{supp}
\newcommand{\fail}{{\mathit{fail}}}

\def\techreport{}

\makeatletter
\RequirePackage[bookmarks,unicode,colorlinks=true]{hyperref}%
   \def\@citecolor{blue}%
   \def\@urlcolor{blue}%
   \def\@linkcolor{blue}%

\def\orcidID#1{\smash{\href{http://orcid.org/#1}{\protect\raisebox{-1.25pt}{\protect\includegraphics{ORCID_Color.eps}}}}}
\makeatother

\usepackage{bbding}

\title{Verified Probabilistic Policies \\ for Deep Reinforcement Learning}

\ifthenelse{\isundefined{\techreport}}{%
\author{Edoardo Bacci \orcidID{0000-0002-0367-898X} \and David Parker \orcidID{0000-0003-4137-8862}  \institute{School of Computer Science, University of Birmingham, UK} \institute{University of Birmingham, Birmingham, United Kingdom \\ \email{\{exb461,d.a.parker\}@bham.ac.uk}}}
}{%
\author{Edoardo Bacci \and David Parker \institute{School of Computer Science, University of Birmingham, UK} \institute{University of Birmingham, Birmingham, United Kingdom \\ \email{\{exb461,d.a.parker\}@bham.ac.uk}}}
}

\begin{document}

\maketitle

\begin{abstract}
Deep reinforcement learning is an increasingly popular technique
for synthesising policies to control an agent's interaction with its environment.
There is also growing interest in formally verifying
that such policies are correct and execute safely. 
Progress has been made in this area by building on existing work for verification of
deep neural networks and of continuous-state dynamical systems.
In this paper, we tackle the problem of verifying \emph{probabilistic} policies
for deep reinforcement learning, which are used to, for example,
tackle adversarial environments, break symmetries and manage trade-offs.
We propose an abstraction approach, based on interval Markov decision processes,
that yields probabilistic guarantees on a policy's execution,
and present techniques to build and solve these models
using abstract interpretation, mixed-integer linear programming,
entropy-based refinement and probabilistic model checking.
We implement our approach and illustrate its effectiveness
on a selection of reinforcement learning benchmarks.
\end{abstract}

\section{Introduction}

Reinforcement learning (RL) is a technique for training a policy used
to govern the interaction between an agent and an environment.
It is based on repeated explorations of the environment,
which yield rewards that the agent should aim to maximise.
\emph{Deep reinforcement learning} combines RL and deep learning,
by using neural networks to store a representation
of a learnt reward function or optimal policy.
These methods have been increasingly successful
across a wide range of challenging application domains, including for example,
autonomous driving~\cite{KendallHJMRALBS19}, robotics~\cite{GHLL17} and healthcare~\cite{YLNY21}.

In safety critical domains, it is particularly important to
assure that policies learnt via RL will be executed safely,
which makes the application of \emph{formal verification} to this problem appealing.
This is challenging, especially for deep RL,
since it requires reasoning about multi-dimensional, continuous state spaces
and complex policies encoded as deep neural networks.

There are several approaches to assuring safety in reinforcement learning,
often leveraging ideas from formal verification,
such as the use of temporal logic to specify safety conditions,
or the use of abstract interpretation to build discretised models.
One approach is \emph{shielding} (e.g.,~\cite{Alshiekh2018}),
which synthesises override mechanisms to prevent
the RL agent from acting upon bad decisions;
another is \emph{constrained} or \emph{safe} RL (e.g.~\cite{FultonP18}),
which generates provably safe policies, typically by restricting
the training process to safe explorations.

An alternative approach, which we take in this paper,
is to verify an RL policy's correctness after it has been learnt,
rather than placing restrictions on the learning process or on its deployment.
Progress has been made in the formal verification of policies for RL~\cite{bastani18}
and also for the specific case of deep RL~\cite{KazakBKS19,BGP21,BP20b},
in the latter case by building on advances in
abstraction and verification techniques for neural networks;
\cite{BGP21} also exploits the development of efficient abstract domains
such as \emph{template polyhedra}~\cite{SankaranarayananSM05},
previously applied to the verification of continuous-space and hybrid systems~\cite{BogomolovFGH17,FrehseGH18}.

A useful tool in reinforcement learning is the notion of a \emph{probabilistic policy}
(or \emph{stochastic policy}), which chooses randomly between available actions in each state,
according to a probability distribution specified by the policy.
This brings a number of advantages
(similarly to mixed strategies~\cite{osborne2004introduction} in game theory
and contextual bandits~\cite{langford2007epoch}), such as
balancing the exploration-exploitation tradeoff~\cite{GF18},
dealing with partial observability of the environment~\cite{papoudakis2021local},
handling multiple objectives~\cite{VDBK09}
or learning continuous actions~\cite{pmlr-v48-mniha16}.

In this paper, we tackle the problem of verifying the safety
of probabilistic policies for deep reinforcement learning.
We define a formal model of their execution using
(continuous-state, finite-branching) \emph{discrete-time Markov processes}.
We then build and solve sound abstractions of these models. 
This approach was also taken in earlier work~\cite{BP20b},
which used Markov decision process abstractions to verify
deep RL policies in which actions may exhibit failures.

However, a particular challenge for probabilistic policies, as generated by deep RL,
is that policies tend to specify very different action distributions across states.
We thus propose a novel abstraction based on \emph{interval Markov decision processes} (IMDPs),
in which transitions are labelled with intervals of probabilities,
representing the range of possible events that can occur.
We solve these IMDPs, over a finite time horizon,
which we show yields \emph{probabilistic guarantees},
in the form of upper bounds on the actual probability of the RL policy
leading the agent to a state designated to be unsafe.

We present methods to construct IMDP abstractions using template polyhedra
as an abstract domain, and mixed-integer linear programming (MILP) to
reason symbolically about the neural network policy encoding
and a model of the RL agent's environment.
We extend existing MILP-based methods for neural networks
to cope with the softmax encoding used for probabilistic policies.
Naive approaches to constructing these IMDPs yield abstractions
that are too coarse, i.e., where the probability intervals are too wide
and the resulting safety probability bounds are too high be useful.
So, we present an iterative refinement approach based on sampling
which splits abstract states via cross-entropy minimisation
based on the uncertainty of the over-approximation.

We implement our techniques, building on an extension of the
probabilistic model checker PRISM~\cite{KNP11} to solve IMDPs.
We show that our approach successfully verifies probabilistic policies
trained for several reinforcement learning benchmarks
and explore trade-offs in precision and computational efficiency.

\startpara{Related work}
As discussed above, other approaches to assuring safety in reinforcement learning include
shielding~\cite{Alshiekh2018,Bastani2019,Zhu2019,Konighofer2020,Jansen2020}
and constrained or safe~RL \cite{FultonP18,HasanbeigAK19,Cheng19,Srinivasan2020,HasanbeigAK20,Ma2021,Jin2021,HFM+21}.
By contrast, we verify policies independently,
without limiting the training process or imposing constraints on execution.

Formal verification of RL, but in a \emph{non-probabilistic} setting includes:
\cite{bastani18}, which extracts and analyses decision trees;
\cite{KazakBKS19}, which checks safety and liveness properties for deep RL;
and \cite{BGP21}, which also uses template polyhedra and MILP to build abstractions,
but to check (non-probabilistic) safety invariants.

In the \emph{probabilistic} setting, perhaps closest is our earlier work~\cite{BP20b},
which uses abstraction for finite-horizon probabilistic verification of deep RL,
but for non-probabilistic policies, thus using a simpler (MDP) abstraction,
as well as a coarser (interval) abstract domain and a different, more basic approach to refinement.
Another approach to generating formal probabilistic guarantees is \cite{DNP22}, which,
unlike us, does not need a model of the environment and
instead learns an approximation and produces
probably approximately correct (PAC) guarantees. 
Probabilistic verification of neural network policies on partially observable models,
but for \emph{discrete} state spaces, was considered in~\cite{CJT21}. 

There is also a body of work on verifying continuous space probabilistic models
and stochastic hybrid systems, by building finite-state abstractions
as, e.g., interval Markov chains~\cite{LAB15} 
or interval MDPs~\cite{LWDA18,CLL+19},
but these do not consider control policies encoded as neural networks.
Similarly, abstractions of discrete-state probabilistic models
use similar ideas to our approach, notably via the use of
interval Markov chains~\cite{FLW06} and stochastic games~\cite{KKNP10}.

\section{Background}\label{sec:bg}

We first provide background on the two key probabilistic models used in this paper:
\emph{discrete-time Markov processes} (DTMPs), used to model RL policy executions,
and \emph{interval Markov decision processes} (IMDPs), used for abstractions.

\startpara{Notation}
We write  $\Dist(X)$ for the set of discrete probability distributions over a set $X$,
i.e., functions $\mu:X\to [0,1]$ where $\sum_{x\in X}\mu(x)=1$.
The support of $\mu$, denoted $\supp(\mu)$,
is defined as $\supp(\mu)=\{x\in X\,|\,\mu(x)>0\}$.
We use the same notation where $X$ is uncountable but where $\mu$ has finite support.
We write $\mathcal{P}(X)$ to denote the powerset of $X$
and $v^i$ for the $i$th element of a vector $v$.



\begin{definition}[Discrete-time Markov process]\label{def:dtmp}
	A (finite-branching) \emph{discrete-time Markov process}
	is a tuple $(S,S_0,\matr{P},\AP,L)$, where:
	$S$ is a (possibly uncountably infinite) set of states;
	$S_0\subseteq S$ is a set of initial states;
	$\matr{P}:S\times S\to [0,1]$ is a transition probability matrix,
	where $\sum_{s'\in \supp(\matr{P}(s,\cdot))}\matr{P}(s,s')=1$ for all $s\in S$;
	$\AP$ is a set of atomic propositions; and $L:S\to\mathcal{P}(\AP)$ is a labelling function.
\end{definition}

A DTMP begins in some initial state $s_0\in S_0$
and then moves between states at discrete time steps.
From state $s$, the probability of making a transition to state $s'$ is $\matr{P}(s,s')$.
Note that, although the state space of DTMPs used here is continuous,
each state only has a finite number of possible successors.
This is always true for our models (where transitions represent
policies choosing between a finite number of actions) and simplifies the model.

A \emph{path} through a DTMP is an infinite sequence of states
$s_0 s_1 s_2\dots$ such that 
$\matr{P}(s_i,s_{i+1})>0$ for all $i$.
The set of all paths starting in state $s$ is denoted $\Path(s)$
and we define a probability space $\Pr{s}{}$ over $\Path(s)$ in the usual way~\cite{KSK76}.
We use atomic propositions (from the set $\AP$) to label states of interest
for verification, e.g., to denote them as safe or unsafe.
For $b\in\AP$, we write $s\models b$ if $b\in L(s)$.

The probability of reaching a $b$-labelled state from $s$ within $k$ steps is:
$$
\Pr{s}{}(\Diamond^{\leq k}\mathit{b})
= \Pr{s}{}(\{s_0 s_1 s_2\dots \in\Path(s)\,|\,s_i\models\mathit{b}\mbox{ for some } 0\leq i \leq k\})
$$
which, since DTMPs are finite-branching models,
can be computed recursively: 
$$
\Pr{s}{}(\Diamond^{\leq k}\mathit{b})=
\left\{\begin{array}{cl}
1 & \mbox{if } s \models\mathit{b} \\
0 & \mbox{if } s \not\models\mathit{b} \land k{=}0 \\
\sum_{s'\in \supp(\matr{P}(s,\cdot))}\matr{P}(s,s')\cdot\Pr{s'}{}(\Diamond^{\leq k-1}\mathit{b}) \ \ & \mbox{otherwise.}
\end{array}\right.
$$



\noindent
To build abstractions, we use interval Markov decision processes (IMDPs).


\begin{definition}[Interval Markov decision process]
	An \emph{interval Markov decision process}
	is a tuple $(S,S_0,\matr{P},\AP,L)$, where:
	$S$ is a finite set of states;
	$S_0\subseteq S$ are initial states;
	$\matr{P}:S\times \Nset\times S\to (\Iset \cup {0})$ is the interval transition probability function,
	where $\Iset$ is the set of probability intervals $\Iset = \{[a,b] \ | \ 0< a \leq b \leq 1\}$,
	assigning either a probability interval or the probability exactly 0 to any transition;
	$\AP$ is a set of atomic propositions; and $L{:}S{\to}\mathcal{P}(\AP)$ is a labelling function.
\end{definition}

Like a DTMP, an IMDP evolves through states in a 
state space $S$, starting from an initial state $s_0\in S_0$.
In each state $s\in S$, an action $j$ must be chosen.
Because of the way we use IMDPs, and to avoid confusion with the actions taken by RL policies,
we simply use integer indices $j\in\Nset$ for actions.
The probability of moving to each successor state $s'$
then falls within the interval $\matr{P}(s,j,s')$.


To reason about IMDPs, we use \emph{policies},
which resolve the nondeterminism in terms of actions and probabilities.
A policy $\sigma$ of the IMDP selects the choice to take in each state, based on the history of its execution so far.
In addition, we have a so-called \emph{environment policy} $\tau$ which selects probabilities for each transition 
that fall within the specified intervals.
For a policy $\sigma$ and environment policy $\tau$,
we have a probability space $\Pr{s}{\sigma,\tau}$
over the set of infinite paths starting in state $s$.
As above, we can define, for example, the probability
$\Pr{s}{\sigma,\tau}(\Diamond^{\leq k}\mathit{b})$ of reaching
a $b$-labelled state from $s$ within $k$ steps, under $\sigma$ and $\tau$.

If $\psi$ is an event of interest defined by a measurable set of paths (e.g., $\Diamond^{\leq k}\mathit{b}$),
we can compute (through \emph{robust value iteration}~\cite{WTM12})
lower and upper bounds on, e.g., maximum probabilities,
over the set of all allowable probability values:
$$\Pr{s}{\max\min}(\psi) = \sup_\sigma \inf_\tau \Pr{s}{\sigma,\tau}(\psi)
\ \ \ \mbox{ and } \ \ \ 
\Pr{s}{\max,\max}(\psi) = \sup_\sigma \sup_\tau \Pr{s}{\sigma,\tau}(\psi)
$$

\section{Modelling and Abstraction of Reinforcement Learning}
\label{sec:abstr_prob}

We begin by giving a formal definition of our model for the execution of a
reinforcement learning system, under the control of a probabilistic policy.
We also define the problem of verifying that this policy is executed safely,
namely that the probability of visiting an unsafe system state,
within a specified time horizon, is below an acceptable threshold.

Then we define abstractions of these models,
given an abstract domain over the states of the model,
and show how an analysis of the resulting abstraction
yields probabilistic guarantees in the form of sound upper
bounds on the probability of a failure occurring.
In this section, we make no particular assumption about the
representation of the policy, nor about the abstract domain.

\subsection{Modelling and Verification of Reinforcement Learning}

Our model takes the form of a controlled dynamical system
over a continuous $n$-dimensional state space $S\subseteq \Rset^n$,
assuming a finite set of \emph{actions} $\Act$ performed at discrete time steps.
A (time invariant) \emph{environment} $E:S\times\Act\rightarrow S$
describes the effect of executing an action in a state,
i.e., if $s_t$ is the state at time $t$ and $a_t$ is the action taken
in that state, we have $s_{t+1} = E(s_t,a_t)$.

We assume a reinforcement learning system is controlled by a
\emph{probabilistic policy}, i.e., a function of the form
$\pi:S\rightarrow\Dist(A)$, where $\pi(s)(a)$ specifies the probability
with which action $a$ should be taken in state $s$.
Since we are interested in verifying the behaviour of a particular policy,
not in the problem of learning such a policy, we ignore issues of partial observability.
We also do not need to include any definition of rewards.

Furthermore, since our primary interest here is in the treatment of probabilistic policies,
we do not consider other sources of stochasticity,
such as the agent's perception of its state
or the environment's response to an action. 
Our model could easily be extended with other discrete probabilistic aspects,
such as the policy execution failure models considered in~\cite{BP20b}.

Combining all of the above, we define an \emph{RL execution model}
as a (continuous-space, finite-branching) \emph{discrete-time Markov process} (DTMP).
In addition to a particular environment $E$ and policy $\pi$,
we also specify a set $S_0\subseteq S$ of possible \emph{initial states}
and a set $S_\fail\subseteq S$ of \emph{failure states},
representing \emph{unsafe} states.

\begin{definition}[RL execution model]\label{def:ctrlmodelprob}
	Assuming a state space $S\subseteq\Rset^n$ and action set $\Act$,
	and given an environment $E:S\times\Act\rightarrow S$,
	policy $\pi:S\rightarrow\Dist(A)$,
	initial states $S_0\subseteq S$ and failure states $S_\fail\subseteq S$,
	the corresponding \emph{RL execution model} 
	is the DTMP $(S,S_0,\matr{P},\AP,L)$ where
	$AP=\{\fail\}$, for any $s\in S$, $\fail\in L(s)$ iff $s\in S_\fail$ and,
	for states $s,s'\in S$:
	$$
	\matr{P}(s,s') = \sum\left\{ \pi(s)(a) \ | \ a\in\Act \mbox{ s.t. } E(s,a)=s' \right\}.
	$$
\end{definition}
The summation in \defref{def:ctrlmodelprob} is required since distinct actions
$a$ and $a'$ applied in state $s$ could result in the same successor state $s'$.

Then, assuming the model above,
we define the problem of verifying that an RL policy executes safely.
We consider a fixed time horizon $k\in\Nset$
and an error probability threshold $p_\mathit{safe}$,
and the check that the probability of reaching an unsafe state
within $k$ time steps is always (from any start state) below $p_\mathit{safe}$.

\begin{definition}[RL verification problem]\label{def:rlverif}
Given a DTMP model of an RL execution, as in \defref{def:ctrlmodelprob},
a time horizon $k\in\Nset$ and a threshold $p_\mathit{safe}\in[0,1]$,
the \emph{RL verification problem} is to check that
$\Pr{s}{}(\Diamond^{\leq k}\mathit{fail})\leq p_\mathit{safe}$
for all $s\in S_0$.
\end{definition}

In practice, we often tackle a \emph{numerical} version of the verification problem,
and instead compute the worst-case probability of error for any start state
$p^+ = \inf\{ \Pr{s}{}(\Diamond^{\leq k}\mathit{fail}) \ | \ s\in S_0\}$
or (as we do later) an upper bound on this value. 

\subsection{Abstractions for Verification of Reinforcement Learning}\label{sec:maths}

Because our models of RL systems are over continuous state spaces,
in order to verify them in practice, we construct finite \emph{abstractions}.
These represent an over-approximation of the original model,
by grouping states with similar behaviour into \emph{abstract states},
belonging to some abstract domain $\hat{S}\subseteq\mathcal{P}(S)$.

Such abstractions are usually necessarily nondeterministic
since an abstract state groups states with similar, but distinct, behaviour.
For example, abstraction of a probabilistic model such as a discrete-time Markov process
could be captured as a Markov decision process~\cite{BP20b}.
However, a further source of complexity for abstracting \emph{probabilistic policies},
especially those represented as deep neural networks,
is that states can also vary widely with regards to the probabilities
with which policies select actions in those states.

So, in this work we represent abstractions as \emph{interval MDPs} (IMDPs),
in which transitions are labelled with intervals,
representing a range of different possible probabilities.
We will show that solving the IMDP (i.e., computing the maximum
finite-horizon probability of reaching a failure state)
yields an upper bound on the corresponding probability
for the model being abstracted.

Below, we define this abstraction and state its correctness,
first focusing separately on abstractions of an
RL system's environment and policy,
and then combining these into a single IMDP abstraction.



\vskip0.7em
Assuming an abstract domain $\hat{S}\subseteq\mathcal{P}(S)$,
we first require an \emph{environment abstraction} $\hat{E}:\hat{S}\times\Act\to\hat{S}$,
which soundly over-approximates the RL environment $E:S\times\Act\rightarrow S$, as follows.


\begin{definition}[Environment abstraction]\label{def:envabs}
For environment $E:S\times\Act\to S$ and set of abstract states $\hat{S}\subseteq\mathcal{P}(S)$,
an \emph{environment abstraction} is a function $\hat{E}:\hat{S}\times\Act\to\hat{S}$ such that:
for any abstract state $\hat{s}\in\hat{S}$, concrete state $s\in\hat{s}$ and action $\act\in\Act$,
we have $E(s,\act) \in \hat{E}(\hat{s},\act)$.
\end{definition}

Additionally, we need, for any RL policy $\pi$, a \emph{policy abstraction} $\hat{\pi}$,
which gives a lower and upper bound on the probability with which
each action is selected within the states grouped by each abstract state.

\begin{definition}[Policy abstraction]\label{dev:policyabs}
    For a policy $\pi:S\rightarrow\Dist(A)$ and a set of abstract states $\hat{S}\subseteq\mathcal{P}(S)$,
    a \emph{policy abstraction} is a pair $(\hat{\pi}_L,\hat{\pi}_U)$
    of functions of the form $\hat{\pi}_L:\hat{S}\times\Act\to [0,1]$
    and $\hat{\pi}_U:\hat{S}\times\Act\to [0,1]$, satisfying the following:
    for any abstract state $\hat{s}\in\hat{S}$, concrete state $s\in\hat{s}$ and action $a\in A$, we have
	$
	\hat{\pi}_L(\hat{s},\act) \leq \pi(s,\act) \leq \hat{\pi}_U(\hat{s},\act)
	$.
\end{definition}


Finally, combining these notions, we can define an \emph{RL execution abstraction},
which is an IMDP abstraction of the execution of an policy in an environment.

\begin{definition}[RL execution abstraction]\label{def:ctrlabsprob}
Let $E$ and $\pi$ be an RL environment and policy,
DTMP $(S,S_0,\matr{P},\AP,L)$ be the corresponding RL execution model
and $\hat{S}\subseteq\mathcal{P}(S)$ be a set of abstract states.
Given also a policy abstraction $\hat{\pi}$ of $\pi$
and an environment abstraction $\hat{E}$ of $E$,
an \emph{RL execution abstraction} is an IMDP
$(\hat{S},\hat{S}_0,\hat{\matr{P}},\AP,\hat{L})$ satisfying the following:
\begin{itemize}
\item for all $s\in S_0$, $s\in\hat{s}$ for some $\hat{s}\in \hat{S}_0$;
\item for each $\hat{s}\in\hat{S}$, there is a partition $\{\hat{s}_1,\dots,\hat{s}_m\}$ of $\hat{s}$
such that, for each $j\in\{1,\dots,m\}$ we have
$\hat{\matr{P}}(\hat{s},j,\hat{s}') = [\hat{\matr{P}}_L(\hat{s},j,\hat{s}'), \hat{\matr{P}}_U(\hat{s},j,\hat{s}')]$
where:
$$
\begin{array}{rcl}
\hat{\matr{P}}_L(\hat{s},j,\hat{s}') & = & \sum\left\{ \hat{\pi}_L(\hat{s}_j,a) \ | \ a\in\Act \mbox{ s.t. } \hat{E}(\hat{s}_j,a)=\hat{s}' \right\} \\
\hat{\matr{P}}_U(\hat{s},j,\hat{s}') & = & \sum\left\{ \hat{\pi}_U(\hat{s}_j,a) \ | \ a\in\Act \mbox{ s.t. } \hat{E}(\hat{s}_j,a)=\hat{s}' \right\}
\end{array}
$$
\item $\AP=\{\fail\}$ and $\fail\in \hat{L}(\hat{s})$ iff $\fail\in L(s)$ for some $s\in\hat{s}$.
\end{itemize}
\end{definition}
Intuitively, each abstract state $\hat{s}$
is partitioned into groups of states $\hat{s}_j$
that behave the same under the specified environment and policy abstractions.
The nondeterministic choice between actions $j\in\{1,\dots,m\}$ in abstract state $\hat{s}$,
each of which corresponds to the state subset $\hat{s}_j$,
allows the abstraction to overapproximate the behaviour of the original DTMP model.

Finally, we state the correctness of the abstraction,
i.e., that solving the IMDP provides upper bounds on the
probability of policy execution resulting in a failure. 
This is formalised as follows (see the appendix for a proof).

\begin{theorem}\label{thm:abstrprob}
Given a state $s\in S$ of an RL execution model DTMP,
and an abstract state $\hat{s}\in \hat{S}$ of the corresponding abstraction IMDP
for which $s\in\hat{s}$: 
$$
\Pr{s}{}(\Diamond^{\leq k}\mathit{fail}) \ \leq \ \Pr{\hat{s}}{\max\max}(\Diamond^{\leq k}\mathit{fail}) .
$$
\end{theorem}
In particular, this means that we can tackle the RL verification problem
of checking that the error probability is below a threshold $p_\mathit{safe}$
for all possible start states (see \defref{def:rlverif}).
We can do this by finding an abstraction for which
$\Pr{\hat{s}}{\max\max}(\Diamond^{\leq k}\mathit{fail}) \leq p_\mathit{safe}$
for all initial abstract states $\hat{s}\in\hat{S}_0$.

Although $\Pr{\hat{s}}{\max\min}(\Diamond^{\leq k}\mathit{fail})$
is not necessarily a \emph{lower} bound on the failure probability,
the value may still be useful to guide abstraction refinement.

\section{Template-based Abstraction of Neural Network Policies}\label{sec:constrabstrprob}

We now describe in more detail the process for
constructing an IMDP abstraction, as given in \defref{def:ctrlabsprob},
to verify the execution of an agent with its environment,
under the control of a probabilistic policy.
We assume that the policy is encoded in neural network form
and has already been learnt, prior to verification,
and we use template polyhedra to represent abstract states.

%
The overall process works by building a $k$-step unfolding of the IMDP,
starting from a set of initial states $\hat{S}_0\subseteq S$.
For each abstract state $\hat{s}$ explored during this process,
we need to split $\hat{s}$ into an appropriate partition $\{\hat{s}_1,\dots,\hat{s}_m\}$.
Then, for each $\hat{s}_j \in \hat{s}$ and each action $a\in\Act$,
we determine lower and upper bounds on the probabilities with which $a$ is selected in states in $\hat{s}_j$,
i.e., we construct a \emph{policy abstraction} $(\hat{\pi}_L,\hat{\pi}_U)$.
We also find the successor abstract state that results from executing $a$ in $\hat{s}_j$,
i.e., we build an \emph{environment abstraction} $\smash{\hat{E}}$.
Construction of the IMDP then follows directly from~\defref{def:ctrlabsprob}.


In the following sections, we describe our techniques in more detail.
First, we give brief details of the abstract domain used: bounded polyhedra.
Next, we describe how to construct policy abstractions via MILP.
Lastly, we describe how to partition abstract states via \emph{refinement}.
We omit details of the environment abstraction
since we reuse the symbolic post operator over template polyhedra given in~\cite{BGP21},
also performed with MILP.
This supports environments specified as
linear, piecewise linear or non-linear systems defined with polynomial and transcendental functions.
The latter is dealt with using linearisation,
subdividing into small intervals and over-approximating using interval arithmetic.

Further details of the algorithms in this section can be found in~\cite{Bac22}.



\subsection{Bounded Template Polyhedra}

Recall that the state space of our model $S\subseteq\Rset^n$ is over $n$ real-valued variables.
We represent abstract states using \emph{template polyhedra}~\cite{SankaranarayananSM05},
which are convex subsets of $\Rset^n$,
defined by constraints in a finite set of \emph{directions} $\Delta \subset \Real^n$
(in other words, the facets of the polyhedra are normal to the directions in $\Delta$).
We call a fixed set of directions $\Delta \subset \Real^n$ a \emph{template}.

Given a (convex) abstract state $\hat{s}\subseteq\Rset^n$,
a $\Delta$-polyhedron of $\hat{s}$ is defined as
the tightest $\Delta$-polyhedron enclosing $\hat{s}$:
\begin{equation*}
  \cap \{ \{ s \colon \inner{\delta}{s} \leq \sup\{ \inner{\delta}{s} \colon s \in \hat{s}\} \} \colon \delta \in \Delta \},\label{eq:tpoly}
\end{equation*}
where $\inner{\cdot}{\cdot}$ denotes scalar product.
In this paper, we restrict our attention to \textit{bounded} template polyhedra (also called \textit{polytopes}),
in which every variable in the state space is bounded by a direction of the template,
since this is needed for our refinement scheme.

Important special cases of template polyhedra 
are \emph{rectangles} (i.e., intervals) and \emph{octagons}.
Later, in \sectref{sec:expts}, we will present an empirical comparison of these different
abstract domains applied to our setting,
and show the benefits of the more general case of template polyhedra.



\subsection{Constructing Policy Abstractions}

We focus first on the abstraction of the RL policy $\pi:S\rightarrow\Dist(A)$,
assuming there are $k$ actions:\ $\Act=\{a_1,\dots,a_k\}$.
Let $\pi$ be encoded by a neural network comprising
$n$ input neurons, $l$ hidden layers, each containing $h_i$ neurons ($1\leq i \leq l$),
and $k$ output neurons, and using ReLU activation functions.

The policy is encoded as follows.
We use variable vectors $z_0\dots,z_{l+1}$ to denote the values of the neurons at each layer.
The current state of the environment is fed to the input layer $z_0$,
each hidden layer's values are as follows:
$$z_i = \mathrm{ReLU}(W_i z_{i-1} + b_i) \mbox{ for } i=1,\dots,l$$
and the output layer is $z_{l+1} = W_{l+1}z_l$,
where each $W_i$ is a matrix of weights connecting layers $i{-}1$ and $i$
and each $b_i$ is a vector of biases.
In the usual fashion, $\mathrm{ReLU}(z) = \max(z,0)$.
Finally, the $k$ output neurons yield the probability assigned by the policy to each action.
More precisely, the probability that the encoded policy selects action $a_j$
is given by $p_j$ based on a softmax normalisation of the output layer:
$$
p_j = \mathrm{softmax}(z_{l+1})^j = \frac{e^{z_{l+1}^{j}}}{\sum^k_{i=1} e^{z_{l+1}^{i}}}
$$
For an abstract state $\hat{s}$, we compute the policy abstraction, i.e., lower and upper bounds
$\hat{\pi}_L(\hat{s},a_j)$ and $\hat{\pi}_U(\hat{s},a_j)$ for all actions $a_j$ (see \defref{dev:policyabs}),
via mixed-integer linear programming (MILP),
building on existing MILP encodings of neural networks~\cite{Tjeng2017,Cheng2017,Bunel2017}.
The probability bounds cannot be directly computed via MILP
due to the nonlinearity of the softmax function so, as a proxy,
we maximise the corresponding entry (the $j$th logit) of the output layer ($l{+}1$).
For the upper bound (the lower bound is computed analogously), we optimise:
\begin{equation}\label{eq:milp}
  \begin{array}{lll}
    \mbox{maximize} & 
    z_{l+1}^j
    \\
    \mbox{subject~to~~~}
    & z_0 \in \hat{s}, \\
    & 0 \leq z_i - W_i z_{i-1} - b_i \leq M z'_i \ \ & \text{for }i = 1,\dots,l,\\
    & 0 \leq z_i \leq M - M z'_i & \text{for }i = 1,\dots,l,\\
    & 0 \leq z'_i \leq 1 & \text{for }i = 1,\dots,l,\\
    & z_{l+1} = W_{l+1}z_l,
  \end{array}
\end{equation}
over the variables $z_0\in\Rset^n$, $z_{l+1}\in\Rset^k$ and $z_i\in\Rset^{h_i}$, $z_i'\in\Zset^{h_i}$ for $1\leq i\leq l$.

Since abstract state $\hat{s}$ is a convex polyhedron, the initial constraint $z_0 \in \hat{s}$
on the vector of values $z_0$ fed to the input layer is represented by $|\Delta|$ linear inequalities.
ReLU functions are modelled using a big-M encoding~\cite{Tjeng2017},
where we add integer variable vectors $z_i'$ and $M\in\Rset$ is a constant
representing an upper bound for the possible values of neurons.

We solve 2$k$ MILPs to obtain lower and upper bounds on the logits for all $k$ actions.
We then calculate bounds on the probabilities of each action by combining these values as described below.
Since the exponential function in softmax is monotonic, it preserves the order of the intervals,
allowing us to compute the bounds on the probabilities achievable in $\hat{s}$.





Let $x_{lb,i}$ and $x_{ub,i}$ denote the lower and upper bounds, respectively,
obtained for each action $a_i$ via MILP
(i.e., the optimised values $z_{l+1}^i$ in (\ref{eq:milp}) above).
Then, the upper bound for the probability of choosing action $a_j$ is $y_{ub,j}$:
$$y_{ub,j} = \mathrm{softmax}(z_{ub,j})
\mbox{\qquad where \qquad}
z_{ub,j}^i = \left\{\begin{array}{cl} x_{ub,i} & \mbox{if } i=j \\ 1-x_{lb,i} \ & \mbox{otherwise}\end{array}\right.$$
and where $z_{ub,j}$ is an intermediate vector of size $k$.
Again, the computation for the lower bound is performed analogously.


\subsection{Refinement of Abstract States}

As discussed above, each abstract state $\hat{s}$ in the IMDP is split into 
a partition $\{\hat{s}_1,\dots,\hat{s}_m\}$ and, for each $\hat{s}_i$,
the probability bounds $\hat{\pi}_{L}(\hat{s}_i,\act)$ and $\hat{\pi}_{U}(\hat{s}_i,\act)$
are determined for each action $\act$.
If these intervals are two wide, the abstraction is too coarse and the results uninformative.
To determine a good partition (i.e., one that groups states with
similar behaviour in terms of the probabilities chosen by the policy),
we use \emph{refinement}, repeatedly splitting $\hat{s}_i$ into finer partitions.

We define the \emph{maximum probability spread} of $\hat{s}_i$, denoted $\Delta^{\max}_{\hat{\pi}}(\hat{s}_i)$, as:
$$\Delta^{\max}_{\hat{\pi}}(\hat{s}_i)=\max_{\act \in \Act} (\hat{\pi}_{U}(\hat{s}_i,\act) - \hat{\pi}_{L}(\hat{s}_i,\act))$$
and we refine $\hat{s}_i$ until $\Delta^{\max}_{\hat{\pi}}(\hat{s}_i)$
falls below a specified threshold $\phi$.
Varying $\phi$ allows us to tune the desired degree of precision.

When refining, our aim is minimise $\Delta^{\max}_{\hat{\pi}}(\hat{s}_i)$,
i.e., to group areas of the state space that have similar probability ranges,
but also to minimise the number of splits performed.
We try to find a good compromise between improving the accuracy of the abstraction
and reducing partition growth, which generates additional abstract states
and increases the size of the IMDP abstraction.


Calculating the range $\Delta^{\max}_{\hat{\pi}}(\hat{s}_i)$
can be done by using MILP to compute each of the lower and upper bounds
$\hat{\pi}_{L}(\hat{s}_i,\act)$ and $\hat{\pi}_{U}(\hat{s}_i,\act)$.
However, this may be time consuming.
So, during the first part of refinement for each abstract state,
we sample probabilities for some states to compute an underestimate of the true range.
If the sampled range is already wide enough to trigger further refinement, we do so;
otherwise we calculate the exact range of probabilities using MILP to check whether there is a need for further refinement.

\begin{figure}[!t]
	\includegraphics[width=\linewidth]{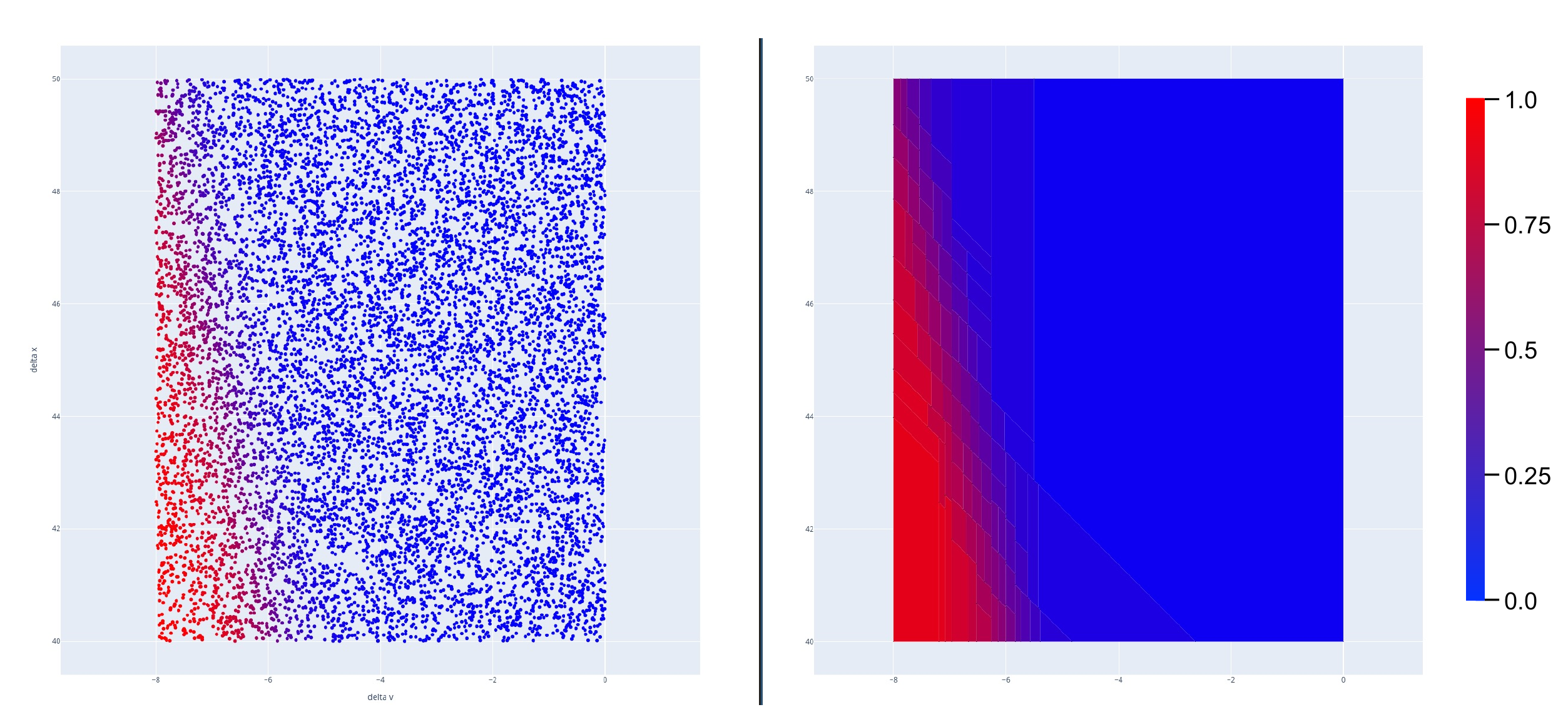}
	\vspace*{-2.3em}
	\caption{Sampled policy probabilities for one action in an abstract state (left)
	and the template polyhedra partition generated through refinement (right).}
	\label{fig:partitioning}
	\vspace*{-1em}
\end{figure}

Each refinement step comprises three phases, described in more detail below:
(i) sampling policy probabilities; (ii) selecting a direction to split; (iii) splitting. 
%
\figref{fig:partitioning} gives an illustrative example of a full refinement.

\startpara{Sampling the neural network policy}
We first generate a sample of the probabilities chosen by the policy within the abstract state.
Since this is a convex region, we sample state points within it randomly using the Hit \& Run method~\cite{hit_and_run}.
We then obtain, from the neural network, the probabilities of picking actions at each sampled state.
We consider each action $a$ separately, and then later split according to the most promising one
(i.e., with the widest probability spread across all actions). 
The probabilities for each $a$ are computed in a \textit{one-vs-all} fashion:
we generate a point cloud representing the probability of taking that action as opposed to any other action.


The number of samples used (and hence the time needed) is kept fixed,
rather than fixing the density of the sampled points. We sample 1000 points per abstract state split but this parameter can be tuned depending on the machine and the desired time/accuracy tradeoff.
This ensures that ever more accurate approximations
are generated as the size of the polyhedra decreases.


\startpara{Choosing candidate directions}
We refine abstract states (represented as template polyhedra)
by bisecting them along a chosen direction from the set~$\Delta$ used to define them.
Since the polyhedra are bounded, we are free to pick any one.
To find the direction that contributes most to reducing the probability spread,
we use cross-entropy minimisation to find the optimal boundary at which to
split each direction, and then pick the direction that yields the lowest value.

Let $\tilde{S}$ be the set of sampled points and $\tilde{Y}_s$ denote the true probability of choosing
action $a$ in each point $s\in\tilde{S}$, as extracted from the probabilistic policy.
For a direction $\delta$, we project all points in $\tilde{S}$ onto $\delta$
and sort them accordingly, i.e., we let $\tilde{S}=\{s_1,\dots,s_m\}$,
where $m=|\tilde{S}|$ and index $i$ is sorted by $\inner{\delta}{s_i}$.
We determine the optimal boundary for splitting in direction $\delta$
by finding the optimal index $k$ that splits $\tilde{S}$ into
$\{s_1,\dots,s_k\}$ and $\{s_{k+1},\dots,s_m\}$.
To do so, we first define the function $Y_i^{k,\delta}$
classifying the $i$th point according to this split:
$$
Y_i^{k,\delta}=
\left\{\begin{array}{cl}
1 & \mbox{if } i \leq k \\
0 & \mbox{if } i > k\\
\end{array}\right.
$$
and then minimise, over $k$, the binary cross entropy loss function:
$$
H(Y^{k,\delta},\tilde{Y}) = -\frac{1}{m}\sum\nolimits_{i=1}^{m}\left(Y_i^{k,\delta} \log(\tilde{Y}_{s_i})+(1-Y_i^{k,\delta})\log(1-\tilde{Y}_{s_i})\right)
$$
which reflects how well the true probability for each point $\tilde{Y}_s$ matches the separation into the two groups.

One problem with this approach is that, if the distribution of probabilities is skewed to strongly favour some probabilities, a good decision boundary may not be picked.
To counter this, we perform sample weighting by grouping the sampled probabilities into small bins,
and counting the number of samples in each bin to calculate how much weight to give to each sample.

\startpara{Abstract state splitting}\label{sec:polyhedron_partitioning}
Once a direction $\delta$ and bisection point $s_k$ are chosen,
the abstract state is split into two with a corresponding pair of
constraints that splits the polyhedron.
Because we are constrained to the directions of the template, and the decision boundary is highly non-linear, sometimes the bisection point falls close to the interval boundary
and the resulting slices are extremely thin.
This would cause the creation of an unnecessarily high number of polyhedra,
which we prevent by imposing a minimum size of the split relative to the dimension chosen. By doing so we are guaranteed a minimum degree of progress and the complex shapes in the non-linear policy space which are not easily classified (such as non-convex shapes) are broken down into more manageable regions.



\section{Experimental Evaluation}\label{sec:expts}

We evaluate our approach by implementing the techniques described in \sectref{sec:constrabstrprob}
and applying them to 3 reinforcement learning benchmarks,
analysing performance and the impact of various configurations and optimisations.



\subsection{Experimental Setup}

\startpara{Implementation}
The code is developed in a mixture of Python and Java.
Neural network manipulation is done through Pytorch~\cite{pytorch},
MILP solution through Gurobi~\cite{gurobi},
graph analysis with {\tt networkX}~\cite{networkx}
and cross-entropy minimisation with Scikit-learn~\cite{scikit-learn}.
IMDPs are constructed and solved using an extension of PRISM~\cite{KNP11}
which implements robust value iteration~\cite{WTM12}.
The code is available from~\url{https://github.com/phate09/SafeDRL}.

\startpara{Benchmarks}
We use the following three RL benchmark environments:

\vskip6pt\noindent
\emph{(i) Bouncing ball}~\cite{JaegerJLLST19}:
The agent controls a ball with height $p$ and vertical velocity $v$,
choosing to either hit the ball downward with a paddle, adding speed, or do nothing.
The ball accelerates while falling and bounces on the ground losing 10\% of its energy;
it eventually stops bouncing if its height is too low and it is out of reach of the paddle.
%
The initial heights and speed vary.
In our experiments, we consider two possible starting regions:
``large'' ($S_0=L$), where $p\in[5,9]$ and $v\in[-1,1]$,
and ``small'' ($S_0=S$), where $p\in[5,9]$ and $v\in[-0.1,0]$.
The safety constraint is that the ball never stops bouncing.


\vskip6pt\noindent
\emph{(ii) Adaptive cruise control}~\cite{BGP21}:
The problem has two vehicles $i\in\{\mathit{lead},\mathit{ego}\}$,
whose state is determined by variables $x_i$ and $v_i$ for the position and speed of each car, respectively.
The lead car proceeds at constant speed (\SI{28}{\meter\per\second}),
and the agent controls the acceleration
($\pm$\SI{1}{\meter\per\second\squared}) of $\mathit{ego}$ using two actions.
The range of possible start states allows a relative distance of $[3,10]$ metres
and the speed of the ego vehicle is in $[26,32]$ m/s.
Safety means preserving $x_\mathit{lead} \geq x_\mathit{ego}$.

\vskip6pt\noindent
\emph{(iii) Inverted pendulum}:
This benchmark is a modified (discrete action) version of
the ``Pendulum-v0" environment from the OpenAI Gym~\cite{Brockman2016}
where an agent applies left or right rotational force to a pole pivoting around one of its ends,
with the aim of balancing the pole in an upright position.
The state is modelled by 2 variables: the angular position and velocity of the pole.
We consider initial conditions of an angle $[-0.05,0.05]$ and speed $[-0.05,0.05]$.
Safety constitutes remaining within a range of positions and velocities
such that an upright position can be recovered.
This benchmark is more challenging than the previous two:
it allows 3 actions (noop, push left, push right)
and the dynamics of the system are highly non-linear, making the problem more complex.

%

\startpara{Policy training}
All agents have been trained using proximal policy optimisation (PPO)~\cite{SchulmanWDRK17} in actor-critic configuration with Adam optimiser. The training is distributed over 8 actors with 10 instances of each environment, managing the collection of results and the update of the network with {\tt RLlib}~\cite{pmlr-v80-liang18b}.
Hyperparameters have been mostly kept unchanged from their default values except the learning rate and batch size which have been set to $5{\times}10^{-4}$ and $4096$, respectively. We used a standard feed forward architecture with 2 hidden layers (size 32 for the bouncing ball and size 64 for the adaptive cruise control and inverted pendulum problems) and ReLU activation functions.


\begin{figure}[!t]
	\begin{subfigure}{.33\textwidth}
		\centering
		\includegraphics[trim={0cm 0cm 2cm 0cm},clip=true,width=3.5cm,height=4cm]{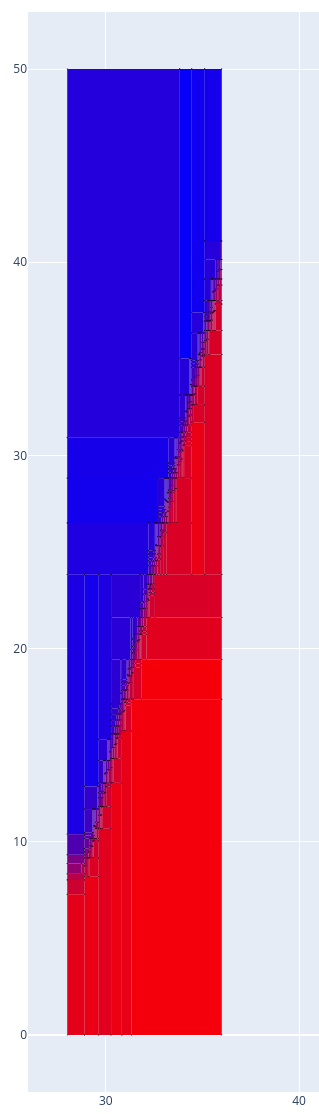}
		\caption{Intervals: $|\hat{s}|=450$}
	\end{subfigure}%
	\begin{subfigure}{.33\textwidth}
		\centering
		\includegraphics[trim={0cm 0cm 2cm 0cm},clip=true,width=3.5cm,height=4cm]{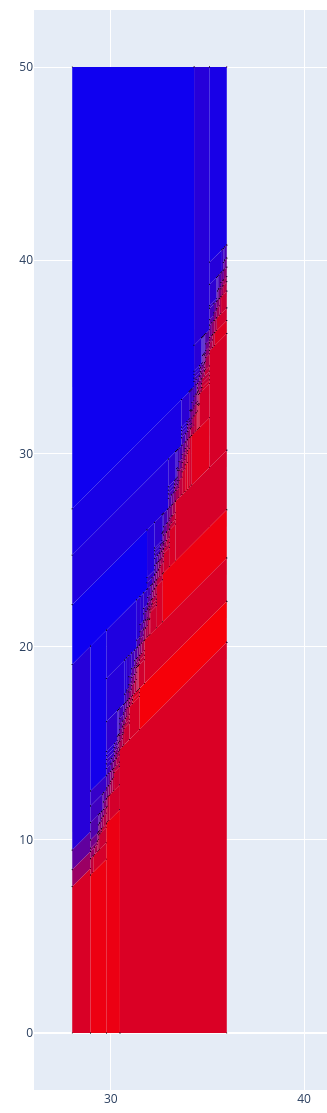}
		\caption{Octagons: $|\hat{s}|=334$}
	\end{subfigure}
	\begin{subfigure}{.33\textwidth}
		\centering
		\includegraphics[trim={0cm 0cm 2cm 0cm},clip=true,width=3.5cm,height=4cm]{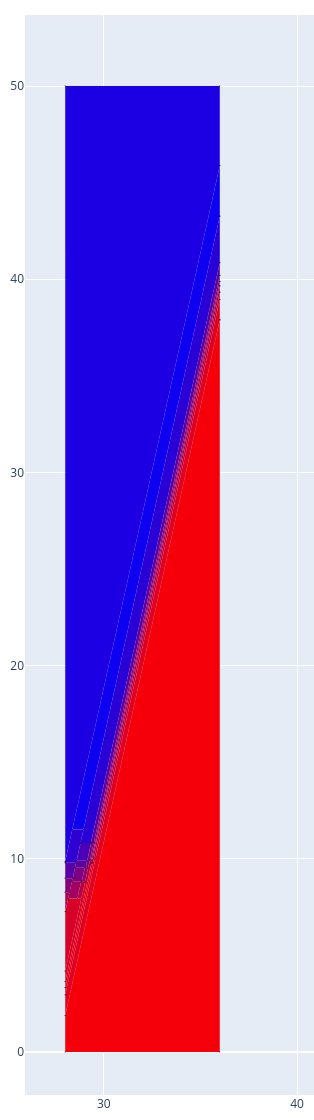}
		\caption{Templates: $|\hat{s}|=25$}
	\end{subfigure}
	\vspace*{-0.2em}
	\caption{Policy abstractions for an abstract state from the adaptive cruise control benchmark, using different abstract domains (see \figref{fig:partitioning} for legend).}
	\label{fig:template_comparison}
	\vspace*{-1em}
\end{figure}

\startpara{Abstract domains}
The abstraction techniques we present in \sectref{sec:constrabstrprob}
are based on the use of template polyhedra as an abstract domain.
As special cases, this includes rectangles (intervals) and octagons.
We use both of these in our experiments, but also the more general case of arbitrary bounded template polyhedra.
In the latter case, we choose a set of directions by sampling
a representative portion of the state space where the agent is expected to operate,
and choosing appropriate slopes for the directions to better represents the decision boundaries.
The effect of the choice of different template can be seen in Fig \ref{fig:template_comparison} where we show a representative abstract state and how the refinement algorithm is affected by the choice of template:
as expected, increasing the generality of the abstract domain results in a smaller number of abstract states.

\startpara{Containment checks}
Lastly, we describe an optimisation implemented for construction of IMDP abstractions,
whose effectiveness we will evaluate in the next section.
When calculating the successors of abstract states to construct an IMDP,
we sometimes find that successors that are partially or fully contained within previously visited abstract states.
Against the possible trade-off of decreasing the accuracy of the abstraction,
we can attempt to reduce the total size of the IMDP that is constructed
by aggregating together states which are fully contained within previously visited abstract states.

\subsection{Experimental Results}

\begin{table}[!t]\centering
	\setlength{\tabcolsep}{3pt}
    \begin{tabular}{l|c|rrc|rrrcc}
		\toprule
		\multicolumn{1}{c|}{Benchmark} &
		\multicolumn{1}{c|}{\multirow{2}{*}{$k$}} &
		\multicolumn{1}{c}{Abs.} &
		\multicolumn{1}{c}{\multirow{2}{*}{$\phi$}} &
		\multicolumn{1}{c|}{Contain.} &
		\multicolumn{1}{c}{Num.} &
		\multicolumn{1}{c}{Num.} &
		\multicolumn{1}{c}{IMDP} &
		\multicolumn{1}{c}{Prob.} &
		\multicolumn{1}{c}{Runtime}
		\\
		\multicolumn{1}{c|}{environment} &
		\multicolumn{1}{c|}{} &
		\multicolumn{1}{c}{dom.} &
		\multicolumn{1}{c}{} &
		\multicolumn{1}{c|}{check} &
		\multicolumn{1}{c}{poly.} &
		\multicolumn{1}{c}{visited} &
		\multicolumn{1}{c}{size} &
		\multicolumn{1}{c}{bound} &
		\multicolumn{1}{c}{(min.)} \\
		\midrule
		\multicolumn{1}{c|}{\multirow{2}{*}{\shortstack[c]{Bouncing \\ ball ($S_0=\textrm{S}$)}}}
		& 20 & Rect & 0.1 & \cmark & 337 & 28 & 411  & 0.0 & 1 \\
		& 20 & Oct & 0.1 & \cmark & 352 & 66 & 484  & 0.0 & 2 \\
		\midrule
		\multicolumn{1}{c|}{\multirow{4}{*}{\shortstack[c]{Bouncing \\ ball ($S_0=\textrm{L}$)}}}
		& 20 & Rect & 0.1 & \cmark & 1727 & 5534 & 7796  & 0.63 & 30 \\
		& 20 & Oct & 0.1 & \cmark & 2489 & 3045 & 6273  & 0.0 & 33 \\
		& 20 & Rect & 0.1 & \xmark & 18890 & 0 & 23337  & 0.006 & 91 \\
		& 20 & Oct & 0.1 & \xmark & 13437 & 0 & 16837  & 0.0 & 111 \\
		\midrule
		\multicolumn{1}{c|}{\multirow{13}{*}{\shortstack[c]{Adaptive \\ cruise \\ control}}}
		& 7 & Rect & 0.33 & \cmark & 1522 & 4770 & 10702 & 0.084 & 85 \\
		& 7 & Oct & 0.33 & \cmark & 1415 & 2299 & 6394& 0.078 & 60 \\
		& 7 & Temp & 0.33 & \cmark & 2440 & 2475 & 9234 & 0.47 & 70 \\
		& 7 & Rect & 0.5 & \cmark & 593 & 1589 & 3776 & 0.62 & 29 \\
		& 7 & Oct & 0.5 & \cmark & 801 & 881 & 3063 & 0.12 & 30 \\
		& 7 & Temp & 0.5 & \cmark & 1102 & 1079 & 4045 & 0.53 & 34 \\
		& 7 & Rect & 0.33 & \xmark & 11334 & 0 & 24184 & 0.040 & 176 \\
		& 7 & Oct & 0.33 & \xmark & 7609 & 0 & 16899 & 0.031 & 152 \\
		& 7 & Temp & 0.33 & \xmark & 6710 & 0 & 14626 & 0.038 & 113 \\
		& 7 & Rect & 0.5 & \xmark & 3981 & 0 & 8395 & 0.17 & 64 \\
		& 7 & Oct & 0.5 & \xmark & 2662 & 0 & 5895 & 0.12 & 52 \\
		& 7 & Temp & 0.5 & \xmark & 2809 & 0 & 6178 & 0.16 & 48 \\
		\midrule
		\multicolumn{1}{c|}{\multirow{2}{*}{\shortstack[c]{Inverted \\ pendulum}}}
		& 6 & Rect & 0.5 & \cmark & 1494 & 3788 & 14726 & 0.057 & 71 \\
		& 6 & Rect & 0.5 & \xmark & 5436 & 0 & 16695 & 0.057 & 69 \\
		\bottomrule
	\end{tabular}
	\vspace*{0.5em}
	\caption{Verification results for the benchmark environments}\label{tab:perf}
	\vspace*{-2em}
\end{table}

\tabref{tab:perf} summarises the experimental results
across the different benchmark environments;
$k$ denotes the time horizon considered.
We use a range of configurations, varying:
the abstract domain used (rectangles, octagons or general template polyhedra);
the maximum probability spread threshold $\phi$
and whether the containment check optimisation is used.

The table lists, for each case:
the number of independent polyhedra generated,
the number of instances in which polyhedra are contained in previously visited abstract states and aggregated together;
the final size of the IMDP abstraction (number of abstract states);
the generated upper bound on the probability of encountering an unsafe state from an initial state;
and the runtime of the whole process.
Experiments were run on a 4-core 4.2 GHz PC with 64 GB RAM. 



Verification successfully produced probability bounds for all environments considered.
Typically, the values of $k$ shown are the largest time horizons we could check,
assuming a 3 hour timeout for verification.
The majority of the runtime is for constructing the abstraction, not solving the IMDP.

As can be seen, the various configurations result in different
safety probability bounds and runtimes for the same environments,
so we are primarily interested in the impact that these choices have
on the trade-off between abstraction precision and performance.
We summarise findings for each benchmark separately.


\startpara{Bouncing ball}
These are the quickest abstractions to construct and verify
due to the low number of variables and the simplicity of the dynamics.
For both initial regions considered, we can actually verify that it is fully safe (maximum probability 0).
However, for the larger one, rectangles (particular with containment checks)
are not accurate enough to show this.

Two main areas of the policy are identified for refinement:
one where it can reach the ball and should hit it
and one where the ball is out of reach and the paddle should not be activated to preserve energy.
But even for threshold $\phi=0.1$ (lower than used for other benchmarks),
rectangular abstractions resulted in large abstract states containing most of the other states visited by the agent,
and which ultimately overlapped with the unsafe region.

\startpara{Adaptive cruise control}
On this benchmark, we use a wider range of configurations.
Firstly, as expected, for smaller values of the maximum probability spread threshold $\phi$,
the probability bound obtained is lower
(the overestimation error from the abstraction decreases, making it closer to the true maximum probability)
but the abstraction size and runtime increase.
Applying the containment check for previously visited states has a similar effect:
it helps reduce the computation time, but at the expense of overapproximation (higher bounds)

The choice of abstract domain also has a significant impact.
Octagons yield more precise results than rectangles, for the same values of $\phi$,
and also produce smaller abstractions (and therefore lower runtime).
On the other hand, general template polyhedra
(chosen to better approximate the decision boundary) 
do not appear to provide an improvement in time or precision on this example,
instead causing higher probability bounds, especially when combined with the containment check.
Our hypothesis is that this abstract domains groups large areas of the state space
(as shown in Fig. \ref{fig:template_comparison}) and this eventually leads to overlaps with the unsafe region.


\startpara{Inverted pendulum}
This benchmark is more challenging and,
while we successfully generate bounds on the probability of unsafe behaviour,
for smaller values of $\phi$ and other abstract domains,
experiments timed out due to the high number of abstract states generated
and the time needed for MILP solution.
The abstract states generated were sufficiently small that 
the containment check could be used to reduce runtime without increasing the probability bound.


\begin{figure}[!t]
	\begin{subfigure}{.5\textwidth}
		\centering
		\includegraphics[width=0.7\linewidth,keepaspectratio]{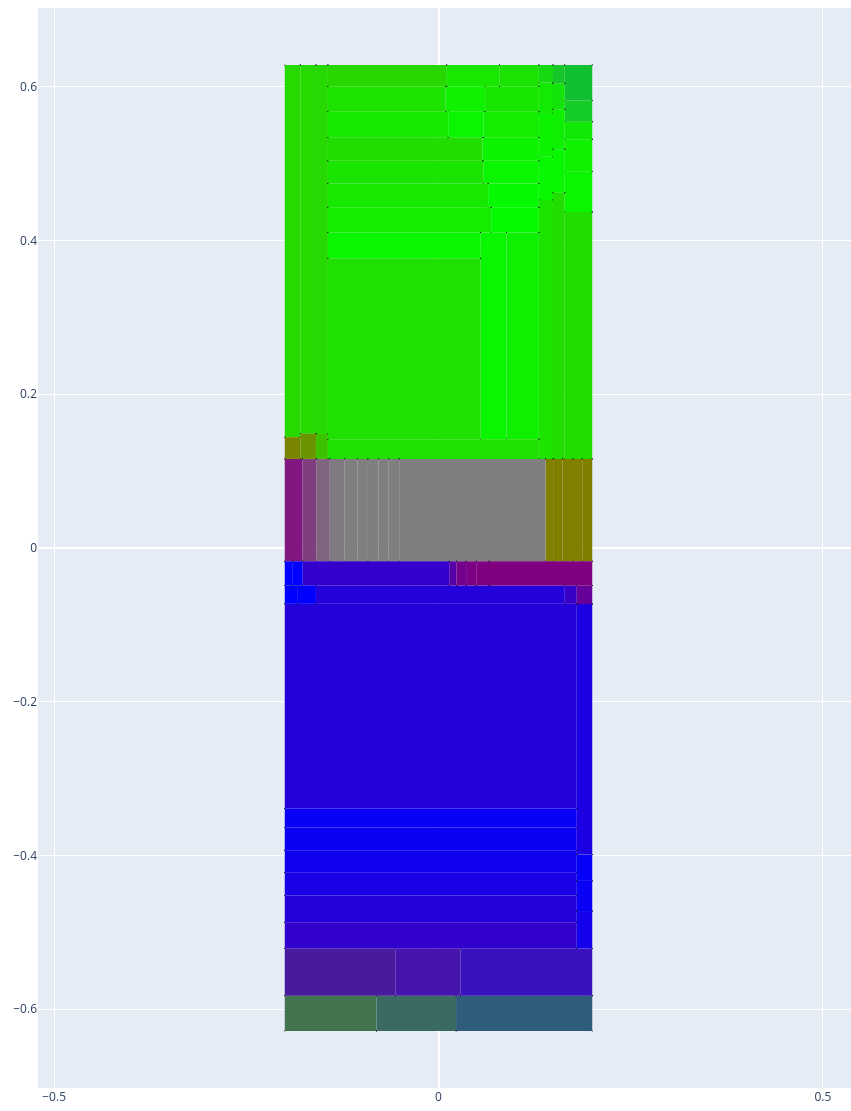}
		\caption{Rectangles}
	\end{subfigure}%
	\begin{subfigure}{.5\textwidth}
		\centering
		\includegraphics[width=0.7\linewidth,keepaspectratio]{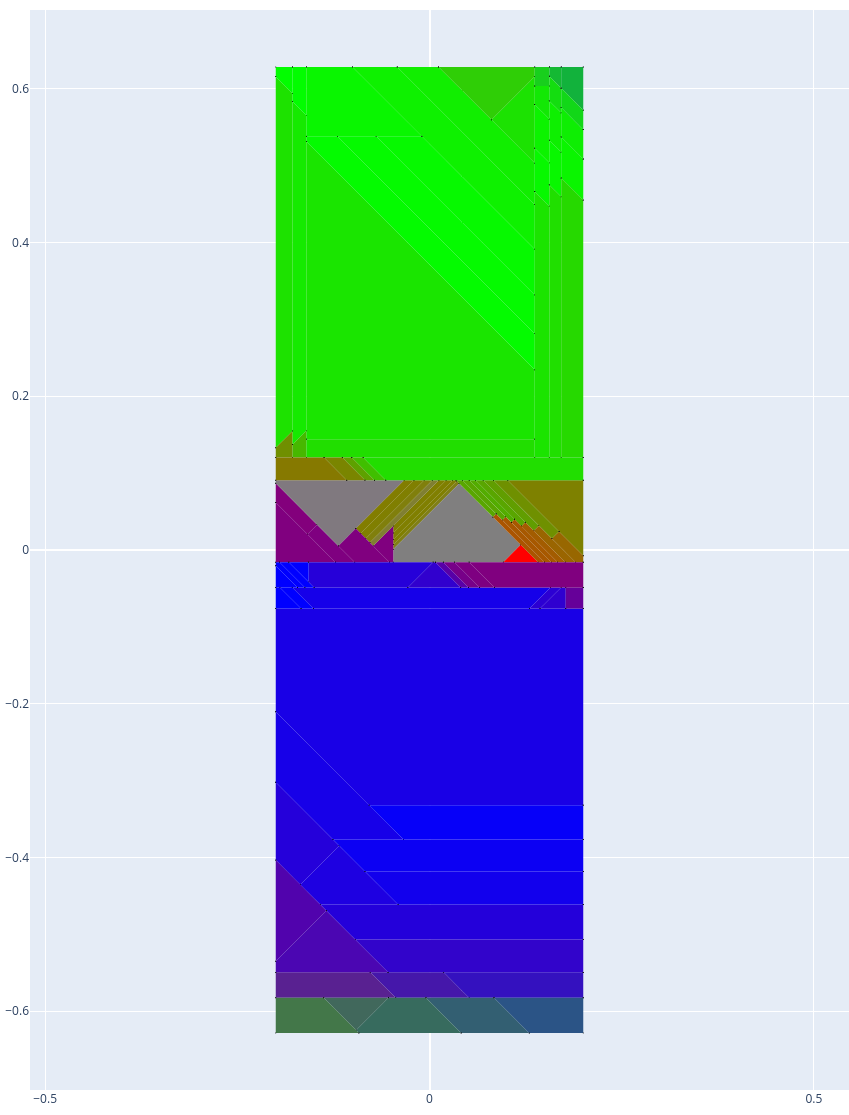}
		\caption{Octagons}
	\end{subfigure}
	\caption{Refined policy abstractions from the inverted pendulum benchmark}
	\label{fig:inverted_pendulum_heatmap}
	\vspace*{-1em}
\end{figure}

\figref{fig:inverted_pendulum_heatmap} illustrates abstraction applied
to a state space fragment from this benchmark using both rectangles and octagons.
It shows the probability of choosing one of three actions,
coded by RGB colour: \emph{noop} (red), \emph{right} (green) and \emph{left} (blue),
The X axis represents angular speed and the Y axis represents the angle of the pendulum in radians.
Notice the grey area towards the centre where all 3 actions have the same probability,
the centre right area with yellow tints (red and green),
and the centre left area with purple tints (red and blue).
Towards the bottom of the heatmap, the colour fades to green
as the agent tries to push the pendulum so that it spins and balances once it reaches the opposite side.

\vspace*{-0.5em}
\section{Conclusion}

We presented an approach for verifying probabilistic policies for deep reinforcement learning agents.
This is based on a formal model of their execution as continuous-space discrete time Markov process,
and a novel abstraction represented as an interval MDP.
We propose techniques to implement this framework with MILP and 
a sampling-based refinement method using cross-entropy minimisation.
Experiments on several RL benchmarks illustrate its effectiveness
and show how we can tune the approach to trade off accuracy and performance.


Future work includes automating the selection of an appropriate template for abstraction
and using lower bounds from the abstraction to improve refinement.

\startpara{Acknowledgements}
This project has received funding from the European Research Council (ERC)
under the European Union’s Horizon 2020 research and innovation programme
(grant agreement No.~834115, FUN2MODEL).


\appendix
\section*{Appendix: Proof of Theorem~\ref{thm:abstrprob}}\label{appx:proof}

We provide here a proof of Theorem~\ref{thm:abstrprob}, from Section~\ref{sec:abstr_prob},
which states that:

\vskip1em

\noindent
Given a state $s\in S$ of an RL execution model DTMP,
and abstract state $\hat{s}\in \hat{S}$ of the corresponding controller abstraction IMDP
for which $s\in\hat{s}$, we have:
$$
\Pr{s}{}(\Diamond^{\leq k}\mathit{fail}) \ \leq \ \Pr{\hat{s}}{\max\max}(\Diamond^{\leq k}\mathit{fail})
$$
By the definition of $\Pr{\hat{s}}{\max\max}(\cdot)$, it suffices to
show that there is \emph{some} policy $\sigma$ and \emph{some} environment policy $\tau$ in the IMDP such that:
\begin{equation}\label{eqn:sigma2}
\Pr{s}{}(\Diamond^{\leq k}\mathit{fail}) \ \leq \ \Pr{\hat{s}}{\sigma,\tau}(\Diamond^{\leq k}\mathit{fail})
\end{equation}
Recall that, in the construction of the IMDP (see Definition~\ref{def:ctrlabsprob}),
an abstract state $\hat{s}$ is associated with a partition of subsets $\hat{s}_j$
of $\hat{s}$, each of which is used to define the $j$-labelled choice in state $\hat{s}$.
Let $\sigma$ be the policy that picks in each state $s$ (regardless of history)
the unique index $j_s$ such that $s\in \hat{s}_{j_s}$.
Then, let $\tau$ be the environment policy that selects the upper bound of the
interval for every transition probability.
We use function $\hat{\matr{P}}_{\tau}$ to denote the chosen probabilities, i.e.,
we have $\hat{\matr{P}}_{\tau}(\hat{s},j_s,\hat{s}') = \hat{\matr{P}}_{U}(\hat{s},j_s,\hat{s}')$ for any $\hat{s},j_s,\hat{s}'$.

The probabilities $\Pr{\hat{s}}{\sigma,\tau}(\Diamond^{\leq k}\mathit{fail})$
for these policies, starting in $\hat{s}$, 
are defined similarly to those for discrete-time Markov processes
(see \sectref{sec:bg}):
\begin{equation*}
\Pr{\hat{s}}{\sigma,\tau}(\Diamond^{\leq k}\mathit{fail})=
\left\{\begin{array}{cl}
1 & \mbox{if } \hat{s}\models\mathit{fail} \\
0 & \mbox{if } \hat{s}\not\models\mathit{fail} \land k{=}0 \\
\sum\limits_{\hat{s}'\in \supp(\hat{\matr{P}}(\hat{s},j_s,\cdot))}\hat{\matr{P}}(\hat{s},j_s,\hat{s}'){\cdot}\Pr{\hat{s}'}{\sigma,\tau}(\Diamond^{\leq k-1}\mathit{fail}) & \mbox{otherwise.}
\end{array}\right.
\end{equation*}
Since this is defined recursively, we prove (\ref{eqn:sigma2}) by induction over $k$.
For the case $k=0$, the definitions of $\Pr{s}{}(\Diamond^{\leq 0}\mathit{fail})$
and $\Pr{\hat{s}}{}(\Diamond^{\leq 0}\mathit{fail})$ are equivalent:
they equal 1 if $s\models\mathit{fail}$ (or $\hat{s}\models\fail$) and 0 otherwise.
From Definition~\ref{def:ctrlabsprob},
$s\models\fail$ implies $\hat{s}\models\fail$.
Therefore, $\Pr{s}{}(\Diamond^{\leq 0}\mathit{fail}) \ \leq \ \Pr{\hat{s}}{\sigma,\tau}(\Diamond^{\leq 0}\mathit{fail})$.

Next, for the inductive step, we will assume, as the inductive hypothesis,
that $\Pr{s'}{}(\Diamond^{\leq k-1}\mathit{fail}) \ \leq \ \Pr{\hat{s}'}{\sigma,\tau}(\Diamond^{\leq k-1}\mathit{fail})$ for $s'\in S$ and $\hat{s}'\in\hat{S}$ with $s'\in \hat{s}'$.
If $\hat{s}\models\fail$ then $\Pr{\hat{s}}{\sigma,\tau}(\Diamond^{\leq k}\mathit{fail})=1 \ \geq \ \Pr{s}{}(\Diamond^{\leq k}\mathit{fail})$.
Otherwise we have:
$$
\begin{array}{rcll}
&& \Pr{\hat{s}}{\sigma,\tau}(\Diamond^{\leq k}\mathit{fail}) \\

& = & \sum_{\hat{s}'\in \supp(\hat{\matr{P}}_{\tau}(\hat{s},j_s,\cdot))}\hat{\matr{P}}_{\tau}(\hat{s},j_s,\hat{s}')\cdot\Pr{\hat{s}'}{}(\Diamond^{\leq k-1}\mathit{fail}) \ \ & \mbox{by defn. of $\sigma$ and $\Pr{\hat{s}}{\sigma,\tau}(\Diamond^{\leq k}\mathit{fail})$} \\

& = & \sum_{\hat{s}'\in \supp(\hat{\matr{P}}_{U}(\hat{s},j_s,\cdot))}\hat{\matr{P}}_{U}(\hat{s},j_s,\hat{s}')\cdot\Pr{\hat{s}'}{}(\Diamond^{\leq k-1}\mathit{fail}) \ \ & \mbox{by defn. of $\tau$} \\

& = & \sum_{a\in\Act}\pi_U(\hat{s},a)\cdot\Pr{\hat{E}(\hat{s}_j,a)}{}(\Diamond^{\leq k-1}\mathit{fail}) \ \ & \mbox{by defn. of $\hat{\matr{P}}_{U}(\hat{s},j,\hat{s}')$} \\

& \geq & \sum_{a\in\Act}\pi(s,a)\cdot\Pr{\hat{E}(\hat{s}_j,a)}{}(\Diamond^{\leq k-1}\mathit{fail}) \ \ & \mbox{since $s\in \hat{s}$ and by Defn.\ref{dev:policyabs}} \\

& \geq & \sum_{a\in\Act}  \pi(s,a)\cdot\Pr{E(s,a)}{}(\Diamond^{\leq k-1}\mathit{fail}) & \mbox{by induction and since, by} \\
&&& \mbox{Defn.~\ref{def:envabs}, $E(s,w)\in\hat{E}(\hat{s}_j,w)$} \\

& = & \sum_{s'\in \supp(\matr{P}(s,\cdot))}\matr{P}(s,s')\cdot\Pr{s'}{}(\Diamond^{\leq k-1}\mathit{fail}) & \mbox{by defn. of $\matr{P}(s,s')$} \\[0.5em]

& = & \Pr{s}{}(\Diamond^{\leq k}\mathit{fail}) & \mbox{by defn. of $\Pr{s}{}(\Diamond^{\leq k}\mathit{fail})$} \\
\end{array}
$$
which completes the proof.

\bibliographystyle{splncs04}
\bibliography{bibliography}

\begin{thebibliography}{10}
\providecommand{\url}[1]{\texttt{#1}}
\providecommand{\urlprefix}{URL }
\providecommand{\doi}[1]{https://doi.org/#1}

\bibitem{Alshiekh2018}
Alshiekh, M., Bloem, R., Ehlers, R., K{\"{o}}nighofer, B., Niekum, S., Topcu,
  U.: {Safe reinforcement learning via shielding}. In: Proc. 32nd AAAI
  Conference on Artificial Intelligence (AAAI'18). pp. 2669--2678 (2018)

\bibitem{Bac22}
Bacci, E.: Formal Verification of Deep Reinforcement Learning Agents. Ph.D.
  thesis, School of Computer Science, University of Birmingham (2022)

\bibitem{BGP21}
Bacci, E., Giacobbe, M., Parker, D.: Verifying reinforcement learning up to
  infinity. In: Proc. 30th International Joint Conference on Artificial
  Intelligence (IJCAI'21). pp. 2154--2160 (2021)

\bibitem{BP20b}
Bacci, E., Parker, D.: Probabilistic guarantees for safe deep reinforcement
  learning. In: Proc. 18th International Conference on Formal Modelling and
  Analysis of Timed Systems (FORMATS'20). LNCS, vol. 12288, pp. 231--248.
  Springer (2020)

\bibitem{Bastani2019}
Bastani, O.: {Safe Reinforcement Learning with Nonlinear Dynamics via Model
  Predictive Shielding}. In: Proceedings of the American Control Conference.
  pp. 3488--3494 (2021)

\bibitem{bastani18}
Bastani, O., Pu, Y., Solar{-}Lezama, A.: Verifiable reinforcement learning via
  policy extraction. In: Proc. 2018 Annual Conference on Neural Information
  Processing Systems (NeurIPS'18). pp. 2499--2509 (2018)

\bibitem{BogomolovFGH17}
Bogomolov, S., Freh\-se, G., Giacobbe, M., Henzinger, T.A.:
  Counterexample-guided refinement of template polyhedra. In: {TACAS} {(1)}.
  pp. 589--606 (2017)

\bibitem{Brockman2016}
Brockman, G., Cheung, V., Pettersson, L., Schneider, J., Schulman, J., Tang,
  J., Zaremba, W.: {OpenAI Gym}  (6 2016)

\bibitem{Bunel2017}
Bunel, R., Turkaslan, I., Torr, P., Kohli, P., Kumar, P.: A unified view of
  piecewise linear neural network verification. In: Proc. 32nd International
  Conference on Neural Information Processing Systems (NIPS'18). pp. 4795--4804
  (2018)

\bibitem{CJT21}
Carr, S., Jansen, N., Topcu, U.: Task-aware verifiable {RNN}-based policies for
  partially observable {Markov} decision processes. Journal of Artificial
  Intelligence Research  \textbf{72},  819--847 (2021)

\bibitem{CLL+19}
Cauchi, N., Laurenti, L., Lahijanian, M., Abate, A., Kwiatkowska, M., Cardelli,
  L.: Efficiency through uncertainty: Scalable formal synthesis for stochastic
  hybrid systems. In: 22nd ACM International Conference on Hybrid Systems:
  Computation and Control (2019)

\bibitem{Cheng2017}
Cheng, C.H., N{\"{u}}hrenberg, G., Ruess, H.: Maximum resilience of artificial
  neural networks. In: Proc. International Symposium on Automated Technology
  for Verification and Analysis (ATVA'17). LNCS, vol. 10482, pp. 251--268
  (2017)

\bibitem{Cheng19}
Cheng, R., Orosz, G., Murray, R.M., Burdick, J.W.: End-to-end safe
  reinforcement learning through barrier functions for safety-critical
  continuous control tasks. In: {AAAI}. pp. 3387--3395. {AAAI} Press (2019)

\bibitem{DNP22}
Delgrange, F., Ann~Now{\,{e}}, G.A.P.: Distillation of {RL} policies with
  formal guarantees via variational abstraction of markov decision processes.
  In: Proc. 36th AAAI Conference on Artificial Intelligence (AAAI'22) (2022)

\bibitem{FLW06}
Fecher, H., Leucker, M., Wolf, V.: Don't know in probabilistic systems. In:
  Valmari, A. (ed.) Proc. SPIN'06. LNCS, vol.~3925, pp. 71--88. Springer (2006)

\bibitem{FrehseGH18}
Frehse, G., Giacobbe, M., Henzinger, T.A.: Space-time interpolants. In: {CAV}
  {(1)}. pp. 468--486. Springer (2018)

\bibitem{FultonP18}
Fulton, N., Platzer, A.: Safe reinforcement learning via formal methods: Toward
  safe control through proof and learning. In: {AAAI}. pp. 6485--6492. {AAAI}
  Press (2018)

\bibitem{GF18}
Garc{\'{i}}a, J., Fern{\'{a}}ndez, F.: Probabilistic policy reuse for safe
  reinforcement learning. ACM Transactions on Autonomous and Adaptive Systems
  \textbf{13}(3),  1–24 (2018)

\bibitem{GHLL17}
Gu, S., Holly, E., Lillicrap, T.P., Levine, S.: Deep reinforcement learning for
  robotic manipulation with asynchronous off-policy updates. In: Proc. 2017
  IEEE International Conference on Robotics and Automation (ICRA'17). pp.
  3389--3396 (2017)

\bibitem{gurobi}
{Gurobi Optimization, LLC}: {Gurobi Optimizer Reference Manual} (2021)

\bibitem{HasanbeigAK19}
Hasanbeig, M., Abate, A., Kroening, D.: Logically-constrained neural fitted
  q-iteration. In: {AAMAS}. pp. 2012--2014. IFAAMAS (2019)

\bibitem{HasanbeigAK20}
Hasanbeig, M., Abate, A., Kroening, D.: Cautious reinforcement learning with
  logical constraints. In: {AAMAS}. pp. 483--491. International Foundation for
  Autonomous Agents and Multiagent Systems (2020)

\bibitem{HFM+21}
Hunt, N., Fulton, N., Magliacane, S., Hoang, T.N., Das, S., Solar-Lezama, A.:
  Verifiably safe exploration for end-to-end reinforcement learning. In: Proc.
  24th International Conference on Hybrid Systems: Computation and Control
  (HSCC'21) (2021)

\bibitem{JaegerJLLST19}
Jaeger, M., Jensen, P.G., Larsen, K.G., Legay, A., Sedwards, S., Taankvist,
  J.H.: Teaching {Stratego} to play ball: Optimal synthesis for continuous
  space {MDPs}. In: {ATVA}. pp. 81--97. Springer (2019)

\bibitem{Jansen2020}
Jansen, N., K{\"{o}}nighofer, B., Junges, S., Serban, A., Bloem, R.: {Safe
  reinforcement learning using probabilistic shields}. In: Proc. 31st
  International Conference on Concurrency Theory (CONCUR'20). vol.~171, pp.
  31--316 (2020)

\bibitem{Jin2021}
Jin, P., Zhang, M., Li, J., Han, L., Wen, X.: {Learning on Abstract Domains: A
  New Approach for Verifiable Guarantee in Reinforcement Learning}  (jun 2021)

\bibitem{KKNP10}
Kattenbelt, M., Kwiatkowska, M., Norman, G., Parker, D.: A game-based
  abstraction-refinement framework for {Markov} decision processes. Formal
  Methods in System Design  \textbf{36}(3),  246--280 (2010)

\bibitem{KazakBKS19}
Kazak, Y., Barrett, C.W., Katz, G., Schapira, M.: Verifying deep-{RL}-driven
  systems. In: Proceedings of the 2019 Workshop on Network Meets {AI} {\&} ML,
  NetAI@SIGCOMM'19. pp. 83--89. {ACM} (2019)

\bibitem{KSK76}
Kemeny, J., Snell, J., Knapp, A.: Denumerable {M}arkov Chains. Springer-Verlag,
  2nd edn. (1976)

\bibitem{KendallHJMRALBS19}
Kendall, A., Hawke, J., Janz, D., Mazur, P., Reda, D., Allen, J., Lam, V.,
  Bewley, A., Shah, A.: Learning to drive in a day. In: {ICRA}. pp. 8248--8254.
  {IEEE} (2019)

\bibitem{Konighofer2020}
K{\"{o}}nighofer, B., Lorber, F., Jansen, N., Bloem, R.: {Shield Synthesis for
  Reinforcement Learning}. In: Lecture Notes in Computer Science (including
  subseries Lecture Notes in Artificial Intelligence and Lecture Notes in
  Bioinformatics). vol. 12476 LNCS, pp. 290--306. Springer, Cham (oct 2020)

\bibitem{KNP11}
Kwiatkowska, M., Norman, G., Parker, D.: {PRISM} 4.0: Verification of
  probabilistic real-time systems. In: Proc. 23rd International Conference on
  Computer Aided Verification (CAV'11). LNCS, vol.~6806, pp. 585--591. Springer
  (2011)

\bibitem{LAB15}
Lahijania, M., Andersson, S.B., Belta, C.: Formal verification and synthesis
  for discrete-time stochastic systems. IEEE Transactions on Automatic Control
  \textbf{60}(8),  2031--2045 (2015)

\bibitem{langford2007epoch}
Langford, J., Zhang, T.: The epoch-greedy algorithm for contextual multi-armed
  bandits. Advances in neural information processing systems  \textbf{20}(1),
  96--1 (2007)

\bibitem{pmlr-v80-liang18b}
Liang, E., Liaw, R., Nishihara, R., Moritz, P., Fox, R., Goldberg, K.,
  Gonzalez, J., Jordan, M., Stoica, I.: {RL}lib: Abstractions for distributed
  reinforcement learning. In: Dy, J., Krause, A. (eds.) Proceedings of the 35th
  International Conference on Machine Learning. Proceedings of Machine Learning
  Research, vol.~80, pp. 3053--3062. PMLR (10--15 Jul 2018)

\bibitem{LWDA18}
Lun, Y.Z., Wheatley, J., D’Innocenzo, A., Abate, A.: Approximate abstractions
  of markov chains with interval decision processes. In: Proc. 6th IFAC
  Conference on Analysis and Design of Hybrid Systems (2018)

\bibitem{Ma2021}
Ma, H., Guan, Y., Li, S.E., Zhang, X., Zheng, S., Chen, J.: {Feasible
  Actor-Critic: Constrained Reinforcement Learning for Ensuring Statewise
  Safety}  (2021)

\bibitem{pmlr-v48-mniha16}
Mnih, V., Badia, A.P., Mirza, M., Graves, A., Lillicrap, T., Harley, T.,
  Silver, D., Kavukcuoglu, K.: Asynchronous methods for deep reinforcement
  learning. In: Balcan, M.F., Weinberger, K.Q. (eds.) Proc. 33rd International
  Conference on Machine Learning. vol.~48, pp. 1928--1937. PMLR (2016)

\bibitem{osborne2004introduction}
Osborne, M.J., et~al.: An introduction to game theory, vol.~3. Oxford
  university press New York (2004)

\bibitem{papoudakis2021local}
Papoudakis, G., Christianos, F., Albrecht, S.V.: Agent modelling under partial
  observability for deep reinforcement learning. In: Proceedings of the Neural
  Information Processing Systems (NeurIPS) (2021)

\bibitem{scikit-learn}
Pedregosa, F., Varoquaux, G., Gramfort, A., Michel, V., Thirion, B., Grisel,
  O., Blondel, M., Prettenhofer, P., Weiss, R., Dubourg, V., Vanderplas, J.,
  Passos, A., Cournapeau, D., Brucher, M., Perrot, M., Duchesnay, E.:
  Scikit-learn: Machine learning in {P}ython. Journal of Machine Learning
  Research  \textbf{12},  2825--2830 (2011)

\bibitem{SankaranarayananSM05}
Sankaranarayanan, S., Sipma, H.B., Manna, Z.: Scalable analysis of linear
  systems using mathematical programming. In: {VMCAI}. pp. 25--41. Springer
  (2005)

\bibitem{SchulmanWDRK17}
Schulman, J., Wolski, F., Dhariwal, P., Radford, A., Klimov, O.: Proximal
  policy optimization algorithms. arXiv:1707.06347  (2017)

\bibitem{hit_and_run}
Smith, R.L.: Efficient {Monte Carlo} procedures for generating points uniformly
  distributed over bounded regions. Operations Research  \textbf{32}(6),
  1296--1308 (1984)

\bibitem{Srinivasan2020}
Srinivasan, K., Eysenbach, B., Ha, S., Tan, J., Finn, C.: {Learning to be Safe:
  Deep RL with a Safety Critic}  (2020)

\bibitem{Tjeng2017}
Tjeng, V., Xiao, K., Tedrake, R.: Evaluating Robustness of Neural Networks with
  Mixed Integer Programming (2017)

\bibitem{VDBK09}
Vamplew, P., Dazeley, R., Barker, E., Kelarev, A.V.: Constructing stochastic
  mixture policies for episodic multiobjective reinforcement learning tasks.
  In: Proc. Australasian Conference on Artificial Intelligence. LNCS,
  vol.~5866, pp. 340--349. Springer (2009)

\bibitem{WTM12}
Wolff, E., Topcu, U., Murray, R.: Robust control of uncertain {Markov} decision
  processes with temporal logic specifications. In: Proc. 51th IEEE Conference
  on Decision and Control (CDC'12). pp. 3372--3379 (2012)

\bibitem{YLNY21}
Yu, C., Liu, J., Nemati, S., Yin, G.: Reinforcement learning in healthcare: A
  survey. ACM Computing Surveys  \textbf{55}(1),  1--36 (2021)

\bibitem{networkx}
Networkx - network analysis in python. \url{https://networkx.github.io/},
  accessed: 2020-05-07

\bibitem{pytorch}
Pytorch. \url{https://pytorch.org/}, accessed: 2020-05-07

\bibitem{Zhu2019}
Zhu, H., Magill, S., Xiong, Z., Jagannathan, S.: {An inductive synthesis
  framework for verifiable reinforcement learning}. In: Proceedings of the ACM
  SIGPLAN Conference on Programming Language Design and Implementation (PLDI).
  pp. 686--701. Association for Computing Machinery (jun 2019)

\end{thebibliography}

\end{document}